# Implementation of an Asymmetric Adjusted Activation Function for Class Imbalance Credit Scoring

Xia Li[a,c,$$], Hanghang Zheng[b,$$], Kunpeng Tao[a,d], Mao Mao[c]

[a] China Development Bank, 100031, Beijing, China

[b] School of Finance, Central University of Finance and Economics, 102206, Beijing, China

[c] University of Cambridge, Cambridge CB2 1TN, United Kingdom

[d] School of Software and Microelctronics, Peking University, 102628, Beijing, China

[$$] Corresponding Authors



**Abstract**

Credit scoring is a systematic approach to evaluate a borrower's probability of default (PD) on a bank loan. The data associated with such scenarios are characteristically imbalanced, complicating binary classification owing to the often-underestimated cost of misclassification during the classifier's learning process. Considering the high imbalance ratio (IR) of these datasets, we introduce an innovative yet straightforward optimized activation function by incorporating an IR-dependent asymmetric adjusted factor embedded Sigmoid activation function (ASIG). The embedding of ASIG makes the sensitive margin of the Sigmoid function auto-adjustable, depending on the imbalance nature of the datasets distributed, thereby giving the activation function an asymmetric characteristic that prevents the underrepresentation of the minority class (positive samples) during the classifier's learning process. The experimental results show that the ASIG-embedded-classifier outperforms traditional classifiers on datasets across wide-ranging IRs in the downstream credit-scoring task. The algorithm also shows robustness and stability, even when the IR is ultra-high. Therefore, the algorithm provides a competitive alternative in the financial industry, especially in credit scoring, possessing the ability to effectively process highly imbalanced distribution data.

## 1. Introduction

**Credit scoring** is a vital component of financial risk management, helping financial institutions evaluate the creditworthiness of clients, both individuals and enterprises, and their ability to repay loans(Bussmann et al., 2021; Hand and Henley, 1997). It estimates the likelihood that a borrower will default on a loan within a certain period following the credit evaluation(Abdou and Pointon, 2011).



Credit scoring is also an important factor in loan and bond pricing(Kriebel and Stitz, 2022; Munkhdalai et al., 2019; Wang et al., 2021), as borrowers with a low probability of default (PD) (low risk) are often offered better interest rates. In retail banking, individual borrowers are evaluated with a score that determines the maximum amount they can borrow from a bank, based on their current credit status. Thus, credit scoring is a scientific, objective, and precise approach for revealing the potential default risk of borrowers and providing valuable information to the financial industry. As data sources become increasingly enriched, artificial intelligence has emerged as a powerful tool for credit scoring. The use of machine-learning (ML) methods to predict customers' potential financial risk has become increasingly popular in recent years. The enriched data sources greatly increase the dimensions that depict a client's history and status, including both daily routine behaviors and financial activities. Combined with complex ML methods, this approach yields convincing rating or scoring outputs, which can significantly boost the decisions of bankers or traders.

ML methods widely used today include typical linear algorithms such as logistic regression(Bensic et al., 2005); support vector machines (SVM)(Bellotti and Crook, 2009; Marceau et al., 2019); k-nearest neighbor (KNN)(Mukid et al., 2018); tree-structured methods like random forest (RF), LightGBM (LGB)(Lextrait, 2023; Wang and Ni, 2020) and XGBoost(Li et al., 2020; Xia et al., 2017); deep-learning (DL) methods including the neural network(West, 2000) and its modified versions(Kriebel and Stitz, 2022; Qian et al., 2023); and methods with attention mechanisms(Kriebel and Stitz, 2022; Tyagi, 2022; Wang et al., 2019). These methods consider various factors in both textual and numeric types and can often predict with high accuracy whether a borrower will default on a loan.

Nevertheless, the main difficulty in credit scoring is the small amount of positive samples(Yu et al., 2018; Zhong and Wang, 2023), leading to a high imbalanced ratio (IR). The positive sample refers to minority customers (those who default), while the negative sample refers to majority customers (those who repay their debt on time)(Shen et al., 2021). Annual reports from listed companies in various stock markets indicate that the default rate of a credit lender is around 0.5% to 2%, corresponding to an IR (IR) of about 50 to 190. The small portion of positive samples can be categorized as a binary classification problem on an imbalanced distributed dataset in ML. In traditional binary classification algorithms, the misclassification cost on the positive and negative samples is considered equally in terms of minimizing the total loss of a classifier during the learning process(Kim and Sohn, 2020). This leads to an overestimated contribution of the majority samples and tends to result in poor classification capacity for the minority class(Crone and Finlay, 2012; Mohammed et al., 2020; Tanha et al., 2020), which is commercially crucial for financial institutions. Thus, solving the imbalanced classification problem is essential in credit scoring.

With the above considerations, recent research has been extensively focused on improving the predictability of various ML techniques on the class-imbalance problem. These methods fall into various categories: data-level,(Chawla et al., 2011), algorithm-level,(Leevy et al., 2018; Li et al., 2018; Tanha et al., 2020), and ensemble-level methods. Data-level methods usually preprocess the dataset by resampling techniques, either by increasing the minority samples or decreasing the majority samples, including random oversampling (ROS)(Batista et al., 2004), random undersampling (RUS)(Batista et



al., 2004), the synthetic minority oversampling technique (SMOTE)(Chawla et al., 2011), and adaptive synthetic technique (ADASYN)(Niu et al., 2020). Algorithm-level methods target modifying method structures(Qian et al., 2023; Tan et al., 2019), activation functions (Munkhdalai et al., 2020; Wang et al., 2021), and cost computation including focal loss(Lin et al., 2017; Liu et al., 2022; Wang et al., 2020), asymmetric loss (ASL)(Ridnik et al., 2021), utilizing of misclassification cost matrix(Wang et al., 2021), and label-distribution-aware margin (LDAM)(Cao et al., 2019), which involves increasing the attention for the minority class(Kim et al., 2019). Ensemble methods relate to the use of ensemble combinations(Nalić et al., 2020) or a combination of data-level and algorithm-level methods(Xiao et al., 2023). Among these methods, the modification of activation functions has become popular. During the learning process, the raw value of the output is mapped to a non-linearity space (for binary classification, it is either positive or negative) after the Sigmoid function. However, owing to the nature of its symmetric curve, the model output after the Sigmoid function may ignore the imbalanced nature of the dataset(Munkhdalai et al., 2020) in credit scoring, thereby underrepresenting the importance of the minority class.

Although extensive research has been focusing on improving the method's recognition and discriminating capabilities by increasing the positive sample weight or using the asymmetric configuration(Mushava and Murray, 2024; Panagiotis Alexandridis et al., 2022; Zhang and Qi, 2024) very few studies, to the best of our knowledge, have simultaneously incorporated dynamic IR reflection and asymmetry into the loss-computation process. However, dynamic adjustment depending on the IR is crucial because the distribution profile of customers differs across financial institutions, which is usually characterized as the proportion of defaulted rate. Therefore, from a practical perspective, it is essential to incorporate the dynamic IR adjustments in algorithmic improvements. To this end, **this paper provides an asymmetric adjusted factor embedded within the Sigmoid activation function (ASIG), which enables the model to dynamically reflect both the dataset's asymmetry in credit scoring and the influence of the IR. The experiments shows the classifiers embedded within the ASIG outperform other traditional classifiers across datasets with a wide-ranging IR values.**

The rest of this paper is organized as follows: **Section 2** provides a brief review of recent advancements in handling imbalanced classification. **Section 3** shows the experimental details and methodology of this study, including a detailed discussion of the development of ASIG. **Section 4** provides a further discussion on the algorithm and compares classifiers using ASIG with traditional classifiers. **Section 5** summarizes the key points and the limitations of this work and provides an outlook on future research.

## 2. Related Works

In credit scoring, binary classification on imbalanced datasets has been a topic of significant interest because the biased nature can impair method performance. Data-level preprocessing targets rebalancing the samples by either increasing the volume of minority samples or decreasing the volume of majority samples. The simplest forms of these methods are random oversampling and random undersampling. SMOTE(Chawla et al., 2011) is a widely used and effective protocol for solving this problem. Recently, generative adversarial networks(Goodfellow et al., 2014) and their modified versions(Kang et al., 2022; Lei et al., 2020; Oreski, 2023; Zhang et al., 2019) have drawn increasing attention in credit scoring



because they offer promising data-augmentation perspectives for rebalancing credit-scoring data. In the meantime, other promising rebalancing techniques have also emerged. Xiao *et al.* conducted extensive experiments on 10 benchmark resampling methods with different classifiers on six commonly used training datasets, with original IRs ranging from 1.2 to 28.1(Xiao et al., 2021). However, oversampling could lead to overfitting by the model(Leevy et al., 2018), while undersampling could cause information loss from the majority class(Arefeen et al., 2020).

Various creative protocols have also been developed at the algorithm level, and initiatives have been undertaken to improve the weight-related factors during loss computation to enhance classification performance for imbalanced data(Liu et al., 2022; Loezer et al., 2020; Madabushi et al., 2020; Ren et al., 2022), focusing on assigning the misclassification costs to positive samples to better fit the skewed distribution(Song et al., 2023). For instance, focal loss was proposed by Lin *et al*. for long-tail object detection and has proven to be an effective modification of the loss function to enhance method performance in binary imbalance classification(Lin et al., 2017). Other works related to LDAM have also been conducted(Cao et al., 2019). Some protocols involve the modification of an activation function, including the Gumbel optimized loss(Panagiotis Alexandridis et al., 2022). In light of the Sigmoid activation function's highly imbalanced data, researchers have replaced it with the generalized-extreme-value (GEV) distribution function to better fit the data distribution(Bridge et al., 2020). The GEV function-embedded convolutional neural network (CNN) outperformed the Sigmoid CNN on highly imbalanced data (with an IR of 50), where the area under the curve (AUC) was largely improved. A study focusing on an unbalanced bioassay dataset systematically showed the influence of changing the loss function to replace classical cross-entropy in binary classifications, showing improvements in both discriminatory power and computational efficiency. In their study, the highest IR reached 2,963(Boldini et al., 2022). Moreover, at the ensemble level, extensive research has been conducted on the design and construction of ML methods tailored to credit-scoring applications. Multimodal methodologies have been employed in scenarios related to credit scoring, as well-documented in research(Babaev et al., 2019; Tavakoli et al., 2023). However, from financial practitioners' practical perspective, little research has focused on designing a mechanism that can incorporate the asymmetric nature of the imbalanced distribution of the client profile datasets, as well as dynamically adjusting the method decision threshold according to the IRs (related to the default rate).

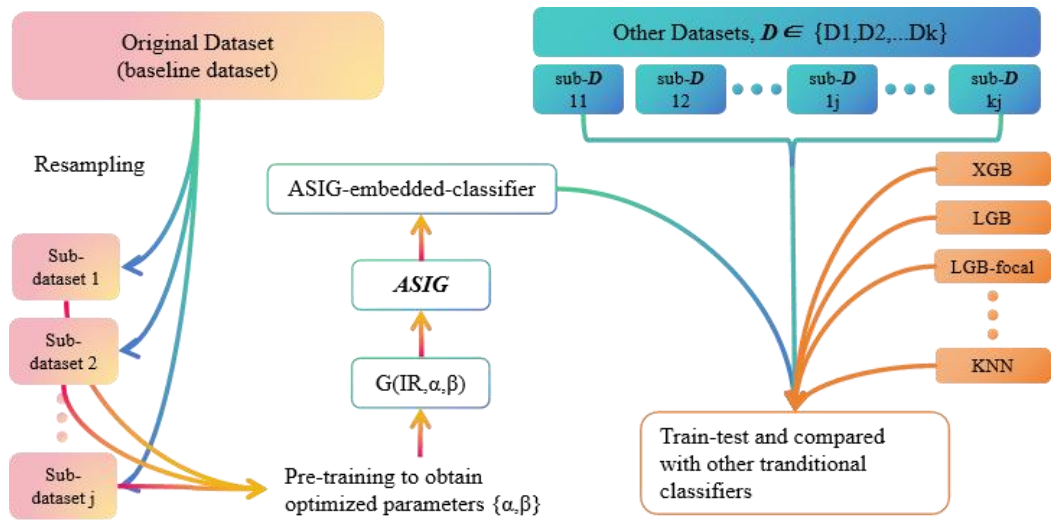



Fig. 1. The general schematic illustration of this study. A baseline dataset is resampled to a series of baseline sub-datasets with various IRs. The adjusted factor *G*, consists of hyper-parameter α, β, depending solely on the IR of the dataset is obtained by pretraining the baseline sub-datasets. The adjusted factor is then embedded in the LightGBM with focal loss configuration, to form the proposed classifier: the ASIG-embedded-classifier. Various classifiers are trained on other datasets, and the performances are compared and analyzed.

To address this problem, which requires considering both the biased nature of the dataset, and the dynamic adjustment of different IRs, we design an ASIG activation function, which is then embedded into the focal-LightGBM classifier, referred to as the ASIG-embedded-classifier, to reflect the asymmetric nature of the clients' distribution in financial datasets. The ASIG-embedded-classifier is IR-dependent, demonstrating a superior prediction performance on datasets across wide-ranging IR values compared with traditional classifiers. Specifically, we engineer a collection of sub-datasets derived from the principal financial credit-scoring datasets. These sub-datasets exhibit IRs from 20 to 300, achieved through a resampling process that mirrors the real-world default rates observed within the credit-scoring sector. Given the exceedingly high IRs in these datasets, we devise a novel activation function, ASIG, by pretraining datasets resampled from a baseline dataset with various IRs. This function incorporates an IR-dependent adjustment factor *G* into the loss-computation process. The introducing of ASIG allows a shift of the Sigmoid curve on the x-axis, thereby imparting an asymmetric quality to the activation function. Such an adjustment is critical to circumvent the underrepresentation of minority classes (positive samples) during the classifier's learning process. Empirical results show that our ASIG method surpasses the performance of LightGBM and other tree-based methods across a broad spectrum of IRs. Furthermore, it exhibits remarkable robustness and stability, even when faced with ultra-high IRs. Our main contributions are as follows:

1. We introduce the ASIG, which embeds an asymmetric adjustment factor into the activation function to reflect both the influence of the biased nature of the dataset, and can dynamically adjust the classifier's threshold depending on the IRs of the dataset;
2. We conduct thorough research on the influence of IRs on the method-classification performance;
3. The proposed ASIG-embedded-classifier's performance, in general, has greater predicting power than baseline classifiers.

## 3. Methodology

This study commences with the resampling of the original datasets sourced from Kaggle, aiming to fabricate sub-datasets spanning diverse IRs. Through initial feature engineering, a suite of sub-datasets, each exhibiting varying IRs, are prepared. A novel IR-dependent adjusted factor, denoted as *G* (IR, α, β), is integrated into the learning algorithm. Subsequently, a pretraining process is carried out, and a baseline dataset is carefully chosen to calibrate and find the hyper-parameters α and β (details can be seen in Section 3.2.2, on the development of ASIG). The adjustment factor *G* is established as solely dependent on the IR and can be used for other datasets. Various classifiers are then trained on these sub-datasets, with their performance, robustness, and stability subjected to an exhaustive comparative analysis. The framework of this study is depicted in **Fig. 1** and **Fig. 2**.

**Framework 1**



| **Framework 1** | |
|---|---|
| **Data**: | *Original Datasets* = (GMC, CHN Banks, Credit Card, Lending Club) |
| **Input**: | Series of sub-datasets with IRs (*IR*$_j$) |
| | Method hyper-parameter $\Theta = (\Theta_0, \Theta_1, ......, \Theta_j)$ |
| **Output** | $\mathcal{D}$ ← Feature engineering for each dataset from (*Original Datasets*) |
| | B$_j$: batches of sub-datasets with different *IR*$_j$ ← randomly resampling ($\mathcal{D}$) |
| **Pretraining**: | Baseline Datasets from resampling of GMC |
| | The $G(IR, \alpha, \beta)$ ← found from pretraining, details to be seen in **Fig. 3** |
| **Output**: | The ASIG-embedded-classifier |
| **Training**: | *for* selected batches with different IR$_j$ from *Original Datasets do* |
| |    Train the methods on sub-datasets |
| | Methods' performances on highly imbalanced datasets are compared |

**Fig. 2.** Delineates the methodological framework of this study. The original datasets, comprising four financial datasets, are found from Kaggle with necessary feature engineering. Then, a compilation of sub-datasets is meticulously constructed via resampling techniques. The hyper-parameter values are then optimized through the pretraining process utilizing the baseline dataset-GMC. Following this optimization, an asymmetric adjusted factor is introduced, which can adjust the original Sigmoid activation function dynamically, and the method performances of the various datasets are analyzed.

## 3.1 Data Preprocessing

In open-source datasets for credit scoring, IR ranges from approximately ~1 to ~200. This reflects a default rate among banking clients of about ~0.5% to ~50%. Such an elevated default rate is incongruent with the majority of real-world credit-lending scenarios. This study concentrates on default ratios that are more representative of actual industry practices, stipulating that the IR of the datasets employed for training and prediction should span from 20 to 300, correlating to default rates ranging from 5% to a mere 0.3%.

| Dataset | Samples | Defaulters | Features | Original IR | IR range after resampling |
|---|---|---|---|---|---|
| GMC | 150000 | 10026 | 11 | 14.0 | [20,300] |
| Credit Card | 30000 | 6630 | 24 | 4.5 | [20,120] |
| CHN Banks | 223868 | 10586 | 37 | 20.2 | [20,300] |
| Lending Club | 14785 | 8305 | 26 | 0.8 | [20,120] |

**Table 1.** Outlines the characteristics of various datasets and their profiles as IRs escalate. Noteworthily, owing to the limited number of defaulters, the Credit Card and Lending Club datasets cannot be resampled to an IR exceeding 120. Exceeding this threshold will result in significant swings in method performance—a phenomenon that will be elaborated upon subsequently.

We select four datasets for method learning and analysis—all sourced from the Kaggle platform. These datasets include GiveMeSomeCredit (GMC), default-of-credit-card-clients (Credit Card, original from UCI dataset), Chinese Banks (CHN Banks), and lending-club (Lending Club, original from the



Financial Services Company Lending Club) datasets. **Table 1** shows the details of the datasets' nature, including the total number of samples, and the IR ranges of each dataset through resampling.

The initial process of our methodology entails the preprocessing of the original datasets through a series of feature engineering steps. This process includes addressing missing values, outliers, categorical data, and the normalization of features to make the data more amenable to diverse algorithmic approaches. Subsequently, we employ random undersampling of the defaulter class to fabricate datasets with different IRs. For the GMC and CHN Banks datasets, the IR is augmented to approximately ~300, at which point the minority class becomes too scant for effective ML application, thereby terminating our dataset construction. Similarly, for the Credit Card and Lending Club datasets, the IR is increased to around 120, beyond which the minority class is deemed insufficient for ML purposes. The datasets are thus finalized with IRs spanning from 20 to 300 for the GMC and CHN Banks datasets, and from 20 to 120 for the Credit Card and Lending Club datasets. We then delve into the alteration of the datasets' distribution by employing a principal component analysis (PCA) for dimensionality reduction. This allows us to visualize the positive and negative classes along the first and second principal components, respectively, represented on the x-axis and y-axis. Utilizing the GMC datasets as a case demonstration, we illustrate the visualized data distributions in **Fig. 3**. It becomes evident that as the IR increases, the prevalence of defaulters—denoted by the blue points—diminishes markedly, signifying the escalating challenge that classifiers face in discerning the distinctive features of the positive class in highly skewed data distributions.

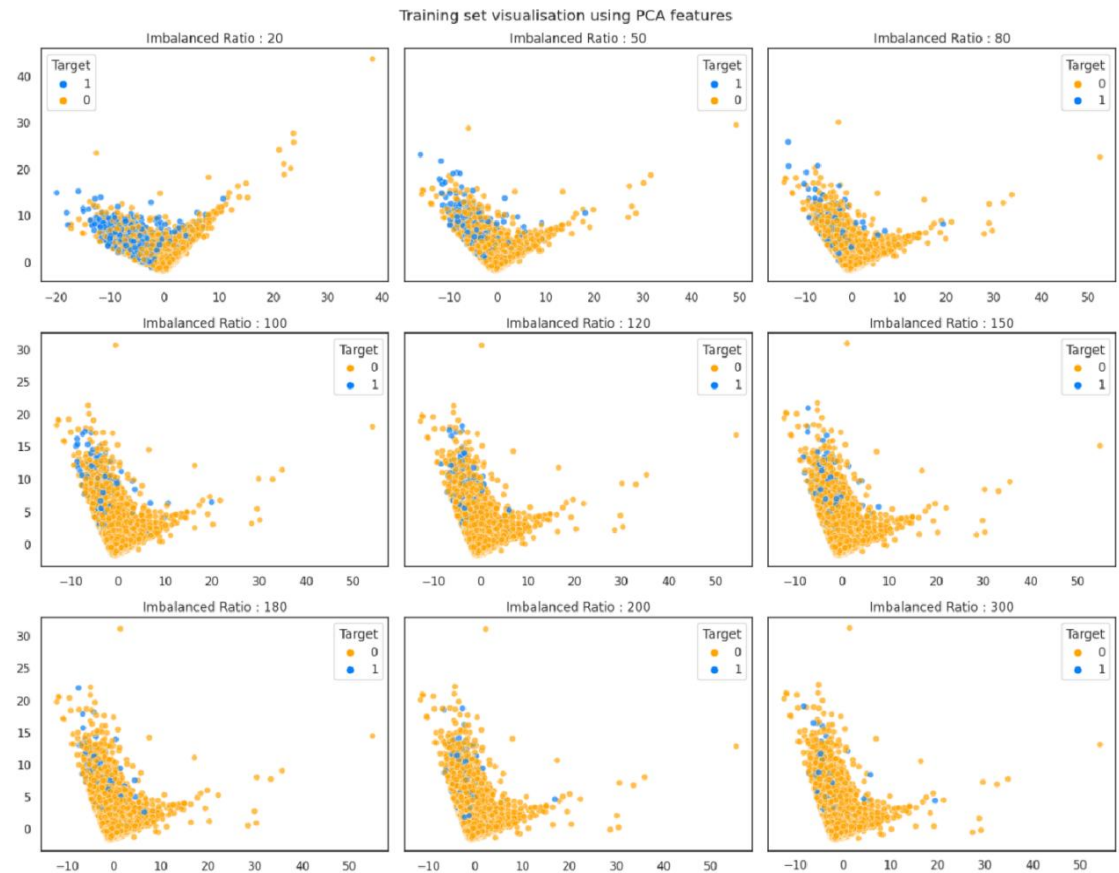



**Fig. 3.** The baseline dataset is presented as an example. The graphical depiction of the processed dataset, delineated across an array of IRs, starkly illustrates the phenomenon that, with the augmentation of the IR, the minority class (positive samples) dwindles in density. This increasing rarity heightens the complexities encountered by analytical methods in their attempt to discern between clients at risk of default and those maintaining fiscal stability.

Financial credit-scoring datasets usually appear in a typical tabular structure. Ensemble tree algorithms including XGBoost (XGB) and LightGBM (LGB) have outperformed linear methods, including logistic regression, simple trees(Fernández-Delgado et al., 2014), and DL methods including FT-Transformer(Grinsztajn et al., 2022), in terms of efficacy and accuracy on the typical tabular dataset. Tree-based ensemble methods also provide enhanced interpretability compared with DL frameworks. Based on the strength of these attributes—simplicity, robustness, and superior performance—we select LGB as our baseline method. Within the LGB architecture, we incorporate the ASIG with the focal loss algorithm—the ASIG-embedded-classifier—to develop a new form for loss computing. We undertake a comparative evaluation of classifier performance, juxtaposing our enhanced method (LGB-Focal with adjustment factor) against a suite of classifiers, including XGB, LGB, LGB-Focal, random forest, decision tree (DT), and KNN. Subsequent sections delve into the intricacies of integrating adjustment factors into the loss computation and carefully evaluate the performance, robustness, and stability of these classifiers amid varying IRs.

## 3.2 The Formulation of ASIG

### 3.2.1 Cost-sensitive Learning Revisited

Cross-entropy (CE) loss is ubiquitously employed as a metric to quantify the misclassification cost between the predicted probabilities and the actual binary labels. The rudimentary binary cross-entropy loss is as follows:

$$L(y, P) = -\frac{1}{N}\sum_{j=1}^{N}(y_j\log(p_j) + (1-y_j)\log(1-p_j)) \qquad (1)$$

where N represents the number of samples; $y_j$ is the ground truth with a binary label for sample j; and $y_j$ = 1 if the clients defaults but is otherwise 0. $P_j$ denotes the predicted probability for sample j, calculated by the Sigmoid function represented as follows:

$$P_j = Sigmoid\ (Z) = \frac{1}{1+exp\ (-Z)} \qquad (2)$$

The Score **Z** from (2) is derived from the matrix weighted product of the feature weights for each sample. The computation of CE loss from (1) signifies that both positive and negative samples have equivalent significance in the cost metric. Given that negative samples constitute the majority of the dataset, they dominate the gradient within the learning process. This predominance is a principal cause in the suboptimal performance often observed in binary classifiers(Tanha et al., 2020) for biased distributed data. The focal loss is engineered to mitigate the challenges by intensifying the learning emphasis on difficult, misclassified instances(Lin et al., 2017). It applies a modulating term to the CE loss to focus learning on hard misclassified examples. Focal loss is defined as follows:



$$FL(y, P) = -\frac{1}{N}\sum_{j=1}^{N} \alpha_j(y_j(1-p_j)^\gamma \log(p_j) + (1-\alpha_j)(1-y_j)(p_j^\gamma \log(1-p_j)) \quad (3)$$

Similarly to CE, here, N denotes the total number of samples, $y_j$ represents the true label, and $p_j$ is the predicted probability for the j$^{th}$ sample. The focal loss thus serves as a refined criterion, and is found to be adept at steering the learning process toward the nuanced detection of imbalanced classes.

### 3.2.2 The Development of ASIG

Prevailing assumptions posit that imbalanced classes adhere to a normal distribution(Yang and Li, 2021), an assumption that may not encapsulate the skewed nature of credit-scoring samples distributions. When evaluating loss via the Sigmoid activation function, it presupposes that the distribution of target samples (the rare events) aligns with a binomial distribution. This is exemplified in (1), where the predicted probability of a positive sample is denoted as $p_j$, and conversely, $1 - p_j$ for the negative. Nonetheless, the distribution of infrequent samples is typically long-tailed and positively skewed, a discrepancy that becomes increasingly significant in methods designed for financial credit loan predictions. Based on the above assumption, we introduce a modified asymmetric element to the Sigmoid function to more accurately represent spaces that are non-normally distributed.

The proposed ASIG is a simple yet effective alternative adjustment for the activation function. Our strategy involves using a pretraining process to find a set of optimized hyper-parameters, $\alpha$ and $\beta$, which are then used to formulate the asymmetric adjusted factor $G \sim (IR, \alpha, \beta)$ to encapsulate the effect of the dataset's imbalanced distribution inside the activation function and to calculate an alternative probability, *Alter_P*, which reflects the biased nature of the dataset by shifting the probability calculated by traditional activation function. The pretraining process is shown in **Fig. 4**.

| **Framework 2 The pretraining process for finding α, β** ||
|---|---|
| **Data**: | ***Baseline Datasets*** ← a series of sub-datasets with various IRs (***IR***) found by resampling the original GMC dataset |
| | ***K***, an array of elements from −3 to 3 |
| **Input**: | Baseline datasets, K |
| **Pretrain**: | *for* each sub-dataset ***L*** with ***IR*** *do* |
| |    *for* each shifted element ***N*** in ***K*** *do* |
| |       Form the asymmetric function ← *Asymmetric Sigmoid* = $\frac{1}{1+exp(-Z+N)}$ |
| |       Embedding the *ASIG* into the focal loss for LightGBM, |
| |       Collect the AUC from each sub-dataset *L* with IR and shifted element N |
| |       Tag the j, k = arg max (AUC(***N***, ***L***)) |
| |    Formulate the mapping of G~*IR*, G←k, IR←j |
| | Log regression of the G, IR in the form G = aln(IR) +b, found α, β ← a,b formulate the ASIG ***G*** = α*Ln(IR) + β |
| **Train** | Other datasets with the built ASIG with the pretrained α, β |

**Fig. 4.** The scheme of the development of ASIG.

The ***G*** functions displace the prototypical Sigmoid curve along the x-axis. The formulation of this



adjustment is delineated as follows:

$$G(IR, \alpha, \beta) = \alpha Ln(IR) + \beta \tag{4}$$

Specifically, we first resample the original baseline dataset (GMC dataset), to synthesize a series of sub-datasets with various IRs ranging from 20 to 200. These synthetic datasets are denoted as the baseline datasets group during the pretraining process. An array of parameters, **K**, is then introduced as the asymmetric element to embed to the Sigmoid activation function, with a searching space $K \in [-3, 3]$ and a step of 0.3, and LightGBM is used as the baseline classifier. The relationship of alternative probability, *Alter_P* (the input value *Z)* and **K**, is expressed in the following form:

$$Alter\_P_{j,k} = ASIG\ (Z)_{j,k} = \frac{1}{1+exp\ (-Z_j+K_k)} \tag{5}$$

where j denotes the sub-dataset j. Different j's values represent datasets with different IR values. The k denotes the index of **K**. The pretraining process undergoes a normal training-testing process, and a series of AUC values are observed on each baseline sub-dataset depending on different **K** values. We set the target as follows:

$$G_j = \arg\max AUC_{j,\ (IR_j)} \tag{6}$$

where $G_j$ is the value of the asymmetric adjusted factor corresponding to the optimal AUC for each sub-dataset. A series of $G_j$ values are then found concerning the IRs, and a mapping of $G \sim (IR, \alpha, \beta)$ is thus constructed. **G** is used as the dependent variable, and the IR is regarded as the independent variable. Given the series D = {($x_1$, $y_1$), ($x_2$, $y_2$)..., ($x_k$, $y_k$)}, ($x_k \in G$, $y_k \in AUC\_series$), it is simple to apply log regression to find the optimized $\alpha$ and β, and thus the asymmetric factor $G$(IR, α, β) is determined. Note that $\alpha$ and β are found through pretraining on sub-datasets derived from the baseline dataset (i.e., the GMC dataset) and are suitable for the use on other datasets and their resampled sub-datasets. This generalization capacity is probably due to the similar structures of the credit-scoring datasets (all are tabular) and their casual inference in nature (as they are all about measuring the credit of the loaners). In this experiment, the calculated values of parameters $\alpha$ and β are 0.24 and 0.92, respectively, with an $R^2$ of 0.689. This pair of values are used through all series of sub-datasets in the following studies. Factoring in the aforementioned adjustments, we can express the proposed variant of the activation function and the ASIG focal loss follows:

$$ASIG\ FL(y, P) = -\frac{1}{N}\sum_{j=1}^{N} \alpha_j(y_j(1-p_j)^\gamma \log(p_j) + (1-\alpha_j)(1-y_j)(p_j^\gamma \log(1-p_j)) \tag{7}$$

$$p = \frac{1}{1+exp\ (-Z+G(Imbalance\ Ratio))} \tag{8}$$

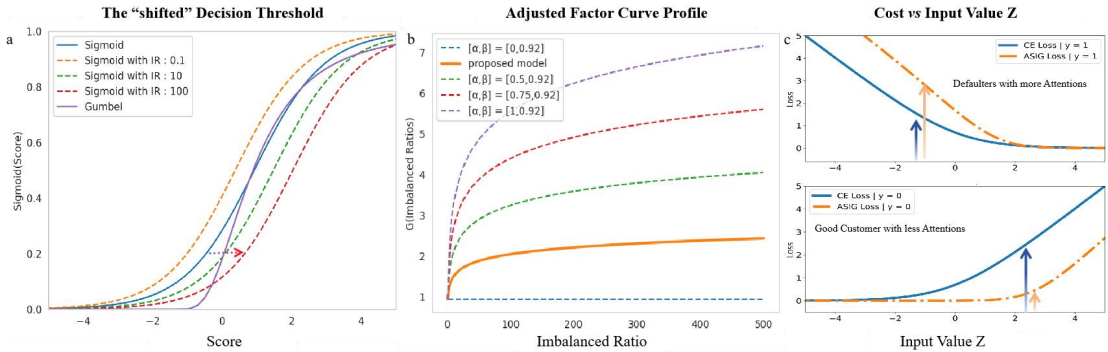

**Fig. 5a:** This figure shows the curve profiles of various activation functions. The Sigmoid function exhibits symmetry with respect to the line x = 0, while the GEV activation function displays inherent asymmetry in its shape. Contrastingly, the modified Sigmoid activation function shows positional asymmetry. Clearly, the Sigmoid curve undergoes a lateral displacement along the x-axis, influenced by the incorporation of the adjusted factor *G*.
**Fig. 5b**: The curve profile of the adjusted factor $G(\text{IR}, \alpha, \beta)$ is depicted with varying pair values of (α) and (β). The curve that has been optimized is highlighted in bold orange. We observe that the upper limit of this curve approaches 2, aligning with the saturation region of the Sigmoid curve, which extends from −2 to 2. **Fig. 5c**: details of the ASIG on the cost of a single sample classification computation process.

In the Discussion section, we present a comparative analysis of the method featuring the ASIG algorithm embedded Focal-LGB against other classifiers across the four datasets. Additionally, we explore the performance variations attributable to differing IRs. This study provides a comprehensive understanding of the method's efficacy in relation to conventional classification approaches in the context of varying data distributions and ultra-high IRs.

## 3.3 Class Imbalance Training

We train various classifiers on the four datasets with wide-ranging IRs. The classifiers applied are the proposed ASIG-embedded-classifier, vanilla LightGBM, LightGBM with focal loss (LGB-Focal), XGBoost (XGB), and several commonly used methods, including random forest, KNN, and decision tree. The trainset size is set at 70% of each dataset. The final AUC scores are the numerical averages of the cross-validation scores.

## 3.4 Method Evaluations

We apply AUC to measure method performance since it is among the most popular indicators for financial banking sectors. AUC is a measure of a binary classifier's ability to distinguish between classes and is used as a summary of the receiver operating characteristic (ROC) curve. In the credit-scoring industry, the actual effect of a prediction method evaluates how much lower the possibility of default is for a good client than for a bad one. In other words, AUC can be interpreted as the probability that a randomly chosen defaulter example is ranked higher than a randomly chosen good customer in this case.

## 4. Discussion

In this section, we first explain the working mechanism of the ASIG-embedded cost function. Compared with traditional CE loss, the ASIG-embedded loss emphasizes the misclassification cost of positive samples while deemphasizing that of negative samples. Subsequently, we show a comparison of the ASIG-embedded-classifier with other classifiers on resampled datasets across wide-ranging IRs. We explore the resilience and consistency of these classifiers by observing the impact of varying IRs. Given the typical scenarios encountered in the financial sector—where the default rate is generally below 5%—we select datasets with IRs ranging from 20 to 300, contingent upon the adequacy of positive samples (defaulters) for the GMC and CHN Banks datasets, and from 20 to 120 in scenarios with a constrained number of defaulters for the CreditCard and LendingClub datasets. The robustness and stability of the proposed ASIG embedded classifier is also discussed.

## 4.1 The ASIG Effects

The asymmetric adjusted factor $G(\text{IR}, \alpha, \beta)$ embedded in ASIG is designed to translocate the Sigmoid curve along the x-axis and thus shift the output probability (by lowering down the decision threshold of the negative samples), as depicted in **Fig. 5a**. The original Sigmoid curve (represented by the blue line) shifts rightward along the x-axis, thereby engendering an asymmetric characteristic and shifting the decision boundary of the classifier. Current studies have shown interest in the introduction of an asymmetric element to the learning process. ASL reflects the asymmetric characteristics of a positive-negative sample imbalance to the loss-computation process(Ridnik et al., 2021) by modifying the cost function. Several research works used the GEV regression to make up for the shortcomings of logistic regression in imbalance classification(Bridge et al., 2020; Calabrese and Osmetti, 2011; Munkhdalai et al., 2020). In comparison, ASIG directly introduces the asymmetric adjustment factor before the calculation of the probability of the positive sample (the output of the Sigmoid function). **Fig. 5b** illustrates a sequence of paired values for $\alpha$ and $\beta$ across wide-ranging IRs. The value of $G$ increases with the IRs, reaching a plateau once the IR has surpassed 200. Given the symmetric nature of the Sigmoid function, with an unsaturated domain spanning around (−2, 2) in **Fig. 5a**, it is inferred that a saturated adjusted factor value of 2 constitutes the boundary. The shift of the symmetric Sigmoid curve can be intuitively interpreted as it lowers the value of probability calculated through an activation function, thus leading the classifiers to pay more attention to the defaulters during the learning process to optimize the total loss during the training process. **Fig. 5c** shows the correlation between the cost and the input value $Z$ for a single sample-prediction process in two cases. When predicting a sample of which the ground truth is negative (a good customer, y = 0, bottom image), the misclassification cost using the ASIG-embedded-classifier is smaller compared with that from CE loss. When predicting a positive sample (a defaulted customer, y = 1, top image), the misclassification cost is emphasized, as the cost calculated from the ASIG-embedded-classifier is higher than that from CE. Emphasizing misclassification cost is an effective way to pay more attention to the positive sample, which has been applied in several loss-function improvements specific to the class-imbalanced scenario. ASIG modifies the decision boundary by adjusting the output of the input value z, thus implementing this mechanism.

The integration of the ASIG confers several key advantages. First, it proactively reduces the probability (the output from the input value $Z$) derived from the activation output, yielding more conservative predictions for positive events and thereby incentivizing classifiers to allocate increased attention to positive samples. Second, it shows a good generalization capacity and computational simplicity. The adjusted factor $G$ is found by pretraining one financial dataset and can be used for other datasets with similar structures (tabular) and downstream scenarios (financial credit scoring). Its incorporation occurs subsequent to the computation of the input value $Z$ and remains invariant throughout the training process. Consequently, it can be considered a constant coefficient for each training set, ensuring computational simplicity.

## 4.2 Method Performance

The GMC dataset serves as the foundational baseline for pretraining the hyper-parameter, specifically for the optimization of α and β. Upon calibrating these parameters, the ASIG method shows superior performance over competing classifiers across datasets with IRs ranging from 20 to 170. The performance details of the methods are listed in **Tables 2,** and the general performance plot is shown in **Fig. 6**. The overall performance of the ASIG method (ASIG-embedded-classifiers), alongside LGB, LGB-focal, and XGB, outperforms that of other classifiers, notably outshining decision tree, KNN, and random forest. Specifically, the ASIG method shows superior performance over LGB and LGB-focal in datasets, with IRs spanning from 20 to 170. However, in datasets characterized by exceedingly high IRs, a pronounced performance degradation is observed at the "breaking point"—a threshold where significant declines vary across different datasets. Additionally, an escalation in variance is noted among classifiers in regions with elevated IRs.

Specifically, as shown in **Fig. 5** and **Table 2**, the proposed classifier outperforms LGB by an average of 1.1%, XGB by 2.2%, LGB-focal by 2.6%, random forest by 6.7%, as well as KNN and decision tree by a substantial 25.3% and 30.1%, respectively. As the IR increases, the ASIG method maintains comparable AUC scores with LGB and LGB-focal. Random forest's performance is approximately 11% worse than its AUC on lower IRs. In scenarios with medium-high IRs (IR < 170), LGB, LGB-focal, XGB, and the ASIG method consistently exhibit robust and high performance, underscoring the algorithms' resilience and their relative insensitivity to distributional shifts in this range. LGB's performance shows a marginal decline as the IR increases, dropping from 0.856 to 0.816, while LGB-focal sees a decrease of 4.3%. The AUC score of the ASIG method decreases from 0.861 to 0.815. The marked performance downturn of LGB and the ASIG method at ultra-high IRs suggests limitations in stability under such conditions. However, they show an acceptable performance decrement at higher IRs, specifically at 230 and 260, indicating a degree of robustness amid increased class disparity.

**AUC of Classifiers**

| IR | ASIG | | LGB | | XGB | | LGB-focal | | RF | | KNN | | DT | |
|---|---|---|---|---|---|---|---|---|---|---|---|---|---|---|
| | | | | | **GMC** | | | | | | | | | |
| 19.7 | **86.1** | ±0.1 | 85.6 | ±1 | 85.3 | ±0.1 | 84.6 | ±0.3 | 83.9 | ±0.1 | 69.7 | ±0.1 | 59 | ±0.2 |
| 38.3 | **85.5** | ±0 | 84.2 | ±0.6 | 84.3 | ±0.2 | 81.2 | ±0.6 | 82 | ±0.5 | 65.3 | ±0.2 | 56.7 | ±0.6 |
| 47.7 | **85.5** | ±0.2 | 84.5 | ±0.2 | 84.4 | ±0.1 | 82.8 | ±0.2 | 81.6 | ±0.1 | 63.9 | ±0.2 | 55.6 | ±0.4 |
| 66.3 | **85.6** | ±0.3 | 84.6 | ±0 | 83.9 | ±1.5 | 82.3 | ±0.7 | 80.4 | ±0.4 | 62.3 | ±0.4 | 54.4 | ±0.1 |
| 94.3 | **85** | ±0.6 | 83.5 | ±0.6 | 82.1 | ±1.7 | 81.5 | ±2.3 | 78.5 | ±0.8 | 59.5 | ±1.2 | 54.2 | ±0.2 |
| 113.0 | **84.4** | ±0.6 | 82.7 | ±1.3 | 82.3 | ±1.4 | 82.7 | ±1.3 | 77.8 | ±0.4 | 56.8 | ±0.8 | 53.2 | ±0.1 |
| 122.3 | **83.9** | ±0.1 | 83.5 | ±0.3 | 82.5 | ±1.3 | 82.9 | ±0.2 | 76.4 | ±0.6 | 57.6 | ±0.7 | 52.5 | ±0.8 |
| 131.7 | **84.2** | ±0.5 | 82.8 | ±0.6 | 82 | ±2 | 81.9 | ±1.4 | 75.3 | ±0.2 | 56.8 | ±0.4 | 51.8 | ±0.3 |
| 141.0 | **86** | ±1.6 | 84.7 | ±1.4 | 83.6 | ±2.2 | 81.7 | ±2.3 | 77.9 | ±0.6 | 57.1 | ±0.3 | 52.5 | ±0.3 |
| 150.2 | **84.4** | ±1.3 | 83.1 | ±0.1 | 79.4 | ±0.3 | 81.6 | ±2 | 74.8 | ±0.3 | 54.9 | ±0.1 | 52.3 | ±0.6 |
| 159.7 | **83.6** | ±1.2 | 82.7 | ±0.4 | 81.3 | ±1 | 82 | ±0.2 | 73.8 | ±0.5 | 54.8 | ±0.1 | 52.3 | ±0.2 |
| 169.0 | **85.4** | ±0.8 | 84.4 | ±0.5 | 82.3 | ±1.8 | 82.4 | ±1 | 76.1 | ±1.8 | 56.4 | ±0 | 52.2 | ±0.1 |

| IR | ASIG | | LGB | | XGB | | LGB-focal | | RF | | KNN | | DT | |
|---|---|---|---|---|---|---|---|---|---|---|---|---|---|---|
| 187.6 | 83.6 | ±0 | 83.9 | ±0.3 | 83.3 | ±0.8 | **84.1** | ±0.4 | 74.7 | ±0.2 | 56 | ±0.2 | 52.5 | ±0.3 |
| 206.2 | **82.7** | ±0.8 | 82.7 | ±0.8 | 81.1 | ±2.9 | 82.3 | ±2 | 73.8 | ±0.4 | 54.8 | ±0.8 | 52.3 | ±0.3 |
| 234.3 | **83.3** | ±0.8 | 82.8 | ±0.3 | 81.5 | ±0.6 | 81 | ±1.5 | 75.5 | ±0.6 | 53.9 | ±0.5 | 52 | ±0.7 |
| 262.2 | 84.4 | ±1.5 | 84.4 | ±2 | 80.8 | ±6.3 | **84.4** | ±2.8 | 75.7 | ±0.6 | 53.8 | ±1.2 | 51.3 | ±1.1 |
| 281.0 | 81.5 | ±0.2 | **81.6** | ±0.3 | 75.5 | ±6.9 | 80.3 | ±2.3 | 72.9 | ±1.9 | 53.7 | ±0.1 | 51.3 | ±0 |
| CHN Banks | | | | | | | | | | | | | | |
| 21.0 | **82.7** | ±0 | 81.9 | ±0.2 | 81.9 | ±0.1 | 81.4 | ±0 | 71.3 | ±0.2 | 60.5 | ±0.2 | 51.3 | ±0 |
| 41.0 | **82.3** | ±0.8 | 81.5 | ±0.9 | 81.4 | ±1.2 | 80.5 | ±2.8 | 70.6 | ±0.8 | 57.1 | ±0.8 | 50.7 | ±0.8 |
| 51.0 | **81.9** | ±0.4 | 81.9 | ±0.4 | 80.4 | ±1.6 | 80 | ±1.6 | 70.4 | ±0.6 | 56.4 | ±0.6 | 51.1 | ±0 |
| 71.0 | **80.3** | ±0.9 | 79 | ±3.6 | 78.3 | ±2.3 | 78.7 | ±3.1 | 68.2 | ±2.1 | 55.2 | ±1.2 | 50.1 | ±0.2 |
| 101.0 | **81** | ±0.7 | 78.2 | ±2.1 | 80.1 | ±2.2 | 79.3 | ±1.3 | 65.9 | ±0.6 | 53.4 | ±0.1 | 50.4 | ±0.1 |
| 121.1 | **81.1** | ±0 | 80.1 | ±0.5 | 80.3 | ±0.1 | 79 | ±2.7 | 65.5 | ±0.3 | 51.6 | ±1 | 50.7 | ±0 |
| 131.0 | **77.6** | ±0.5 | 76.1 | ±0.6 | 76.8 | ±2.5 | 75.7 | ±2.2 | 65.8 | ±1.4 | 52.3 | ±0.1 | 50.8 | ±0.5 |
| 140.8 | **79.8** | ±1.7 | 78 | ±0.1 | 76.5 | ±1.9 | 74.3 | ±1.4 | 61.6 | ±0.8 | 51.1 | ±1.6 | 49.8 | ±0.2 |
| 150.8 | **79.4** | ±0.6 | 75.6 | ±1.1 | 73.2 | ±4.7 | 79.3 | ±2 | 62.4 | ±0.3 | 51 | ±1.3 | 50.2 | ±1 |
| 160.9 | **78.8** | ±1.7 | 75.3 | ±1.3 | 78 | ±1 | 77.1 | ±1 | 61.6 | ±1.7 | 51.2 | ±0.9 | 49.9 | ±0.7 |
| 171.2 | **76.7** | ±0.4 | 71 | ±2.8 | 67 | ±0.5 | 74.6 | ±2 | 60.8 | ±2.1 | 52.1 | ±1.2 | 50 | ±0.3 |
| 180.9 | 79.2 | ±2.1 | 73.8 | ±0 | **79.5** | ±2 | 76.3 | ±0.3 | 64.8 | ±0.6 | 50.8 | ±0.5 | 50.1 | ±0.7 |
| 201.0 | **80.4** | ±1.5 | 72.5 | ±6.1 | 72 | ±0.8 | 78.6 | ±3.6 | 63.5 | ±2.3 | 52.1 | ±0.3 | 50.2 | ±0.3 |
| 221.2 | **76.1** | ±0.8 | 71.6 | ±2.3 | 69.8 | ±3.6 | 68.5 | ±1.3 | 62.9 | ±0.2 | 54 | ±0.8 | 50.5 | ±0.7 |
| 250.7 | **74.9** | ±3.6 | 65.3 | ±3.3 | 71 | ±9.8 | 71.6 | ±2.6 | 60 | ±0.4 | 51.7 | ±0.2 | 49.9 | ±0.4 |

**Table 2.** AUC scores of different classifiers on financial datasets.

Upon the calibration of the hyper-parameters α and β, found by training the GMC and its resampled datasets, subsequent evaluations are conducted on the CHN Banks, Credit Card, and Lending Club datasets without additional adjustments to the α and β values. The sole variable in the ASIG method is the IR, dynamically derived by computing the $G(\text{IR}, \alpha, \beta)$. The performance outcomes of the classifiers across the various datasets are presented in **Fig. 5.** Similarly to the results with the GMC dataset, at relatively low IRs, the ASIG method, along with LGB, LGB-focal, and XGB, markedly outpace the other classifiers. The ASIG method's performance is notably superior to that of LGB, LGB-focal, and XGB. A performance decrement is observed across all classifiers as the IR increases for each dataset, with the rate of decline exhibiting significant variability across the different datasets, as seen in other works(Buda et al., 2018; Korycki and Krawczyk, 2021).

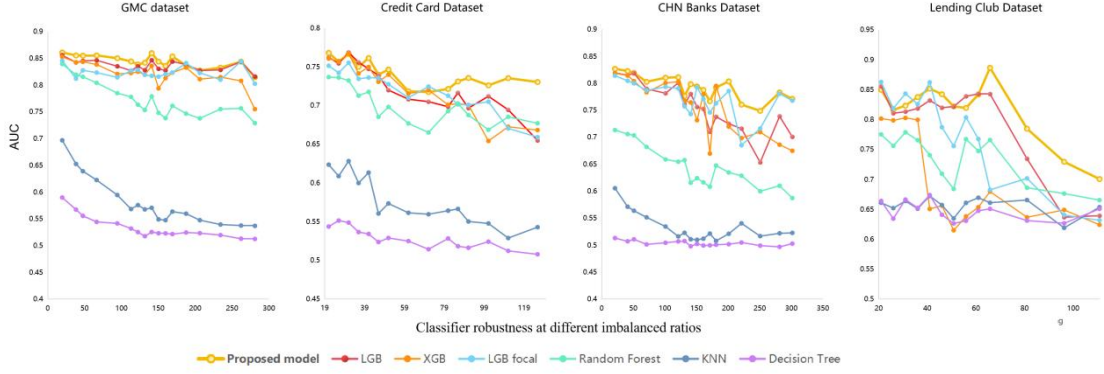

**Fig. 6.** The discriminatory power by **the** AUC scores of different methods' vs IRs (left) and the enlarged view of the focused methods' performances. Dataset1: GMC dataset and datasets with different IRs derived from GMC dataset. Dataset2: Credit Card dataset and datasets with different IRs derived from Credit Card dataset.

We also observe that the performance of all methods exhibits considerable variation with the increase in IR on predicting the Lending Club Dataset (**Table. S1**). A notable decline in the AUC score of XGB is recorded at an IR of around 40, with a similar downturn being seen for LGB-focal at an IR of around 60. Conversely, enhancements in the performance of LGB and the ASIG method are observed at an IR of 70. The dramatic fluctuations are presumed to be a consequence of the limited number of defaulters in the resampled dataset; specifically, only 324 defaulters with an IR of 20 exist in the dataset, and 130 defaulters with an IR of 50 exist in the dataset. This scarcity of defaulters can lead to a lack of consistent characteristic information, thereby affecting the methods' performance.

## 4.3 Robustness and Stability

In addressing class-imbalance issues, it is imperative that robustness and stability are given equal consideration to method performance(Dablain et al., 2022). An ideal classifier not only exhibits high efficacy on datasets characterized by a significant IR but also maintains minimal performance deterioration and consistent variance in the face of escalating IRs(Dablain et al., 2022).

The AUC-ROC curves for each classifier pairing, alongside the resampled datasets (GMC, CHN Banks, and Credit Card datasets), are illustrated in **Fig. 6**. The shaded areas represent the standard deviation of each classifier's performance. As anticipated, the standard deviation expands with the increase in IR. The ASIG method exhibits a distinctly improved performance and reduced variance at elevated IRs when juxtaposed with LGB.

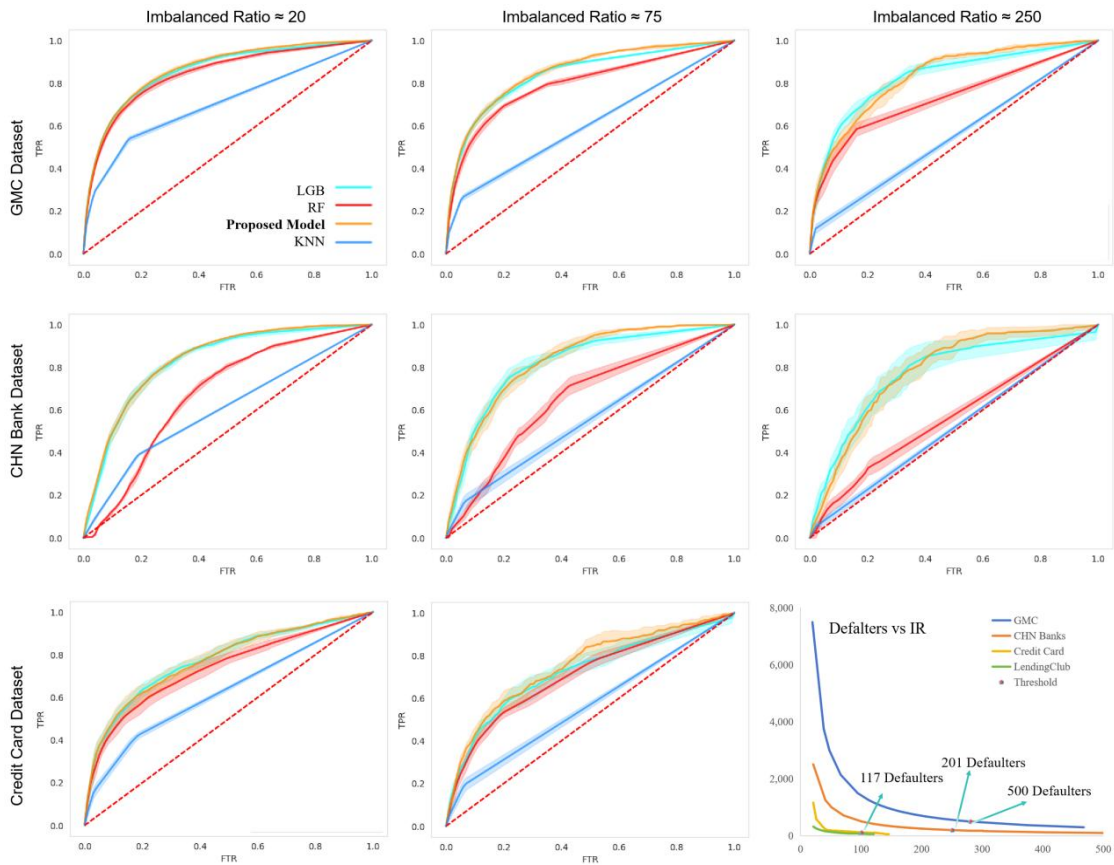

**Fig. 6** The first eight plots present the AUC scores of various methods across datasets with differing IRs. The shaded areas illustrate the variance for each classifier. The KNN classifier exhibits limited predictive ability, while the RF method underperforms across all datasets. The ASIG methods excel, outperforming others in every dataset featured in this figure, and display a notably smaller variance compared with LGB. The ***Defaulter vs IRs*** plot at the bottom right provides a distinct visualization of the positive sample counts (the defaulters) against the changing IRs. The highlighted values denote the "breaking point" at which all classifiers experience a significant decline in performance. This demonstrates that maintaining a defaulter count above 200 is advantageous for achieving robust and stable classifier performance.

The robustness of the method exhibits variability across datasets, even when the IRs are comparable. This robustness appears to be more closely associated with the count of defaulters, as depicted in the Defaulters vs IRs plot in **Fig. 6**. The annotated values on this plot correlate with instances where methods experience performance declines or fluctuates. We observe that when the sample size diminishes to as few as 150 defaulters, regardless of the total sample's IR, the AUC scores begin to fluctuate, and the variance increases. This suggests that such sample volumes are insufficient to provide a stable distribution for the positive samples. The "breaking point"—the juncture at which the method becomes unstable—is identified at 201 defaulters for the CHN Banks dataset and 117 defaulters for the Credit Card dataset. In contrast, the method's performance on the GMC dataset remains relatively stable; no marked decrease is noted even at an ultra-high IR of 300. This stability may be attributed to the fact that, despite the high IR, the distribution of positive samples remains stable enough for method training. Conversely, the Lending Club dataset, with an IR of 20, has only 324 defaulters, which likely contributes to the observed performance swings across all methods.

## 5. Conclusion

In this study, we have put efforts to mitigate the pronounced impact of IRs on the discriminative capacity of methods within the realm of financial credit scoring. We have introduced a simple yet effective activation function, the ASIG, that incorporates an IR-dependent adjusted factor, $G(\text{IR}, \alpha, \beta)$, into the Sigmoid function. This is further embedded in the focal loss and enhanced the performance of classifier. The ASIG-embedded-classifier showed effectiveness and generalization capacity on various datasets with similar structures (tabular), different IRs in financial credit scoring. The integration of the ASIG imparts an asymmetric characteristic to the classifier, thereby counteracting the effects of a highly imbalanced distribution on binary classification challenges.

The ASIG method, alongside a suite of conventional baseline algorithms, undergoes training across various datasets with diverse IRs. As anticipated, a decrease in method performance is observed with the rising IRs. Notably, when the IRs escalate from 20 to 170, the ASIG method consistently outperforms other classifiers across nearly all sub-datasets. For example, it surpasses LGB by an average of 1.1% and XGB by 2.2% in the GMC dataset. Moreover, the ASIG method exhibits robustness and stability within the tested range of IRs, aligning well with the data distributions commonly encountered in financial banking and credit-loan scenarios. Despite the superior performance of the ASIG-embedded-classifier over other classifiers, an increase in IRs leads to fluctuations and augmented variances across all methods.

Given the demonstrated efficacy, resilience, and consistency of the ASIG method on datasets with high IRs, we believe that this algorithm is a viable contender for credit scoring in the financial sector. It is particularly advantageous for applications with highly imbalanced data distributions. Future work will focus on the detailed theoretical mechanisms of the ASIG, a comprehensive study on the effects of pretraining the optimized ASIG hyper-parameter on different datasets, and the engineering implementation of the optimization process for finding improved form and hyper-parameter of the ASIG.

**Abbreviations**

| | |
|---|---|
| ADASYN | Adaptive synthetic |
| ASIG | Asymmetric adjusted factor embedded Sigmoid activation function |
| ASL | Asymmetric loss |
| AUC | Area under the curve. |
| CHN Banks | Chinese banks dataset from Kaggle |
| CNN | Convolutional neural network |
| CreditCard | Credit card default dataset from Kaggle |
| GEV | Generalized extreme value |
| GMC | Give me some credit dataset from Kaggle |
| IR | Imbalanced ratio |
| KNN | K-nearest neighbor |
| LDAM | Label-distribution-aware margin |
| LendingClub | Lending Club dataset, original from the Financial Services Company Lending Club |
| LGB | LightGBM, Light Gradient Boosting Machine |
| PD | Probability of default |
| RF | Random forest |
| ROS | Random oversampling |
| RUS | Random undersampling |
| SMOTE | Synthetic minority oversampling technique |
| SVM | Support vector machine; |
| XGB | XGBoost, eXtreme Gradient Boosting |

# Supplementary Material

**Model Performance Summary**

The details of model performance can be seen in **Table. S1.**, in the predictive analysis of the CHN Banks and Credit Card datasets, the ASIG method exhibits a significant enhancement in performance over traditional classifiers. The AUC scores of the ASIG surpass those of other classifiers for IRs ranging from 20 to 170 within the CHN Banks dataset. This performance improvement is more pronounced as the IRs increases. However, at relatively high IRs, all methods experience considerable performance volatility. This fluctuation can be attributed to the reduced number of positive samples, which may have resulted in the distribution of randomly undersampled datasets failing to provide consistent characteristic information for the positive samples. Taking into account the performance of classifiers on the CHN Banks dataset as an instance, we find that the ASIG method outperforms other classifiers across nearly every different IR. It also outperforms LGB by an average of 3.6%, XGB by 3.9%, LGB-focal by 2.2%, and random forest by 14.9%, as well as KNN and decision tree by a substantial 26.0% and 28.9%, respectively. In scenarios with medium-high IRs (IR < 170), LGB, LGB-focal, XGB, and the ASIG method maintain robust and high performance. However, XGB's performance fluctuates as the IR increases to 180. The AUC scores of the ASIG method decline from 0.827 to 0.771, LGB from 0.813 to 0.700, XGB from 0.819 to 0.675, and LGB-focal from 0.814 to 0.768. It is also important to note that the performance of all classifiers remains relatively stable at IRs below 100 for both the CHN Banks and Credit Card datasets. An IR of 100 corresponds to a default rate of 1%, which is a commonly observed PD in banking.

**AUC of Classifiers**

| IR | ASIG | | LGB | | XGB | | LGB-focal | | RF | | KNN | | DT | |
|---|---|---|---|---|---|---|---|---|---|---|---|---|---|---|
| | | | | | | **Lending Club** | | | | | | | | |
| 21.0 | 85 | ±3.1 | 85.6 | ±2.8 | 80.2 | ±2.1 | **86.3** | ±3.2 | 77.5 | ±5.5 | 66.1 | ±2.4 | 66.4 | ±4.4 |
| 26.0 | 81.6 | ±2.7 | 81.1 | ±3.3 | 79.9 | ±2.7 | **81.9** | ±3.2 | 75.6 | ±2.7 | 65.2 | ±5.2 | 63.4 | ±1.9 |
| 31.0 | 82.3 | ±1.2 | 81.3 | ±2.8 | 80.3 | ±2.9 | **84.3** | ±2.2 | 77.9 | ±2.3 | 66.4 | ±1.3 | 66.6 | ±2.7 |
| 36.0 | **83.8** | ±5 | 81.9 | ±5.6 | 80 | ±2.5 | 82.6 | ±1.1 | 76.6 | ±6.6 | 65.1 | ±4.1 | 65.2 | ±4.2 |
| 41.0 | 85.2 | ±2.8 | 83.2 | ±3.2 | 65.1 | ±5.6 | **86.2** | ±3.5 | 74.1 | ±3.9 | 67.2 | ±4.7 | 67.4 | ±3.4 |
| 46.0 | **84.2** | ±3.7 | 82 | ±5.7 | 65.6 | ±5.3 | 78.7 | ±5.2 | 70.9 | ±5.9 | 65.7 | ±4.3 | 64.1 | ±5.3 |
| 50.9 | **82.2** | ±3.5 | 82.2 | ±5.4 | 61.5 | ±4.9 | 75.6 | ±3.2 | 68.4 | ±9.7 | 63.5 | ±4.8 | 62.6 | ±5 |
| 55.9 | 81.9 | ±8.2 | **83.9** | ±2.4 | 63.8 | ±5.6 | 80.4 | ±10.3 | 76.7 | ±2.1 | 66.1 | ±2.7 | 63.1 | ±4.1 |
| 61.0 | 84.2 | ±6.7 | **84.3** | ±4.4 | 65.4 | ±4.6 | 76.8 | ±4.2 | 74.7 | ±6 | 66.9 | ±4.5 | 64.8 | ±4.3 |
| 65.8 | **88.7** | ±4.1 | 84.3 | ±2.9 | 68 | ±6.2 | 68.3 | ±5.9 | 76.6 | ±5 | 66.1 | ±7.2 | 65.1 | ±7.1 |
| 81.0 | **78.5** | ±5.9 | 73.5 | ±4.3 | 63.7 | ±3.8 | 70.2 | ±8.5 | 68.6 | ±8.5 | 66.6 | ±7.5 | 63.1 | ±3.9 |
| 96.3 | **72.9** | ±10.2 | 63.7 | ±4.1 | 64.9 | ±3.6 | 64 | ±7 | 67.6 | ±8.1 | 61.9 | ±2.9 | 62.7 | ±4.5 |
| 110.8 | **70.1** | ±8.3 | 63.9 | ±6.3 | 62.4 | ±8.2 | 63.2 | ±3.9 | 66.5 | ±11.8 | 65.4 | ±6.2 | 65.1 | ±7.6 |

| IR | ASIG | | LGB | | XGB | | LGB-focal | | RF | | KNN | | DT | |
|---|---|---|---|---|---|---|---|---|---|---|---|---|---|---|
| 121.0 | **69.9** | ±13.8 | 65.3 | ±5 | 52.3 | ±3.8 | 54.4 | ±9.4 | 64.1 | ±4.5 | 58.7 | ±7 | 60.5 | ±6.4 |
| **Credit Card** | | | | | | | | | | | | | | |
| 21.0 | **76.8** | ±0.7 | 76.2 | ±0.9 | 76.1 | ±1.3 | 75.1 | ±1.5 | 73.7 | ±1.4 | 62.4 | ±1.7 | 54.4 | ±1.3 |
| 26.0 | **75.7** | ±2.1 | 75.4 | ±1.3 | 75.7 | ±1.4 | 74.2 | ±1.8 | 73.6 | ±1.9 | 60.9 | ±1.2 | 55.1 | ±1.2 |
| 31.0 | 76.6 | ±2.3 | **76.9** | ±3.2 | 76.7 | ±2.7 | 75.5 | ±2.9 | 73.2 | ±2.7 | 62.8 | ±2.1 | 54.9 | ±1.6 |
| 36.0 | 75 | ±3 | **75.6** | ±1.8 | 74.1 | ±2.3 | 73.5 | ±2.2 | 71.3 | ±2 | 60 | ±2.2 | 53.6 | ±1 |
| 41.0 | **76.1** | ±1.3 | 74.7 | ±2.9 | 75 | ±2.2 | 73.6 | ±1.7 | 71.8 | ±2.8 | 61.3 | ±2.2 | 53.4 | ±2 |
| 46.0 | **74** | ±0.8 | 73.9 | ±2 | 73.1 | ±2.8 | 73.5 | ±2.9 | 68.5 | ±1.6 | 56 | ±1.6 | 52.4 | ±1.5 |
| 51.0 | **74.7** | ±1.2 | 72 | ±4.4 | 74 | ±2.3 | 72.8 | ±2.6 | 69.8 | ±2.1 | 57.3 | ±1.5 | 52.9 | ±1.4 |
| 61.0 | **71.8** | ±3.4 | 70.8 | ±2 | 71.5 | ±4.2 | 71 | ±2.9 | 67.7 | ±1.9 | 56.1 | ±1.2 | 52.5 | ±0.7 |
| 71.1 | 71.7 | ±3.1 | 70.5 | ±1.4 | 72 | ±3.8 | **72.4** | ±1.7 | 66.5 | ±4.6 | 56 | ±1.4 | 51.4 | ±1.3 |
| 80.9 | **72.1** | ±2.7 | 69.8 | ±3.8 | 70.1 | ±2.8 | 71.3 | ±4.7 | 69.3 | ±4 | 56.4 | ±3.8 | 52.8 | ±1.7 |
| 85.9 | **73.1** | ±4.2 | 71.7 | ±4.3 | 70.3 | ±3.3 | 70.1 | ±5.4 | 70.3 | ±5.9 | 56.6 | ±3.6 | 51.8 | ±1.8 |
| 91.1 | **73.6** | ±1.2 | 69.7 | ±4.5 | 70.1 | ±2.8 | 70.1 | ±2.6 | 68.8 | ±3.9 | 55 | ±2.2 | 51.6 | ±1.8 |
| 101.2 | **72.6** | ±6.8 | 71.2 | ±3.7 | 65.4 | ±4.1 | 70.5 | ±3.2 | 66.9 | ±2 | 54.8 | ±1.9 | 52.4 | ±2.1 |
| 111.1 | **73.5** | ±4.4 | 69.4 | ±2.1 | 67.3 | ±5.9 | 67 | ±7.1 | 68.5 | ±8.3 | 52.9 | ±2 | 51.2 | ±1.4 |
| 125.8 | **73** | ±0.9 | 65.5 | ±5.1 | 66.8 | ±4.2 | 65.9 | ±5.2 | 67.7 | ±5.6 | 54.3 | ±2 | 50.8 | ±0.9 |

**Table. S1.** AUC scores of different classifiers on financial datasets.